\documentclass[runningheads]{llncs}
\UseRawInputEncoding
\usepackage{graphicx}
\usepackage{booktabs}
\usepackage{multirow}
\usepackage{adjustbox}
\usepackage{amssymb}
\usepackage{stfloats}
\usepackage{hyperref}
\titlerunning{Generating Prompts through Image-Text Contrastive Learning for VG}

\begin{document}
%
\title{ProGEO: Generating Prompts through Image-Text Contrastive Learning for Visual Geo-localization}
%

\author{Jingqi Hu\inst{1,2} , Chen Mao\inst{1,2}}

\institute{Shanghai Institute of Microsystem and Information Technology, Chinese Academy of Sciences, Shanghai, 200050, China \and
University of Chinese Academy of Sciences, Beijing, 101408, China\\
\email{\{jingqihu, chen.mao\}@mail.sim.ac.cn}}
%
\maketitle              
\begin{abstract}
Visual Geo-localization (VG) refers to the process to identify the location described in query images, which is widely applied in robotics field and computer vision tasks, such as autonomous driving, metaverse, augmented reality, and SLAM. In fine-grained images lacking specific text descriptions, directly applying pure visual methods to represent neighborhood features often leads to the model focusing on overly fine-grained features, unable to fully mine the semantic information in the images. Therefore, we propose a two-stage training method to enhance visual performance and use contrastive learning to mine challenging samples. We first leverage the multi-modal description capability of CLIP (Contrastive Language-Image Pretraining) to create a set of learnable text prompts for each geographic image feature to form vague descriptions. Then, by utilizing dynamic text prompts to assist the training of the image encoder, we enable the image encoder to learn better and more generalizable visual features. This strategy of applying text to purely visual tasks addresses the challenge of using multi-modal models for geographic images, which often suffer from a lack of precise descriptions, making them difficult to utilize widely. We validate the effectiveness of the proposed strategy on several large-scale visual geo-localization datasets, and our method achieves competitive results on multiple visual geo-localization datasets. Our code and model are available at https://github.com/Chain-Mao/ProGEO.

\end{abstract}
\section{Introduction}
\label{sec:intro}
Visual Geo-localization (VG) is a challenging task in the fields of computer vision \cite{Arandjelovic_2018_netvlad,Torii_2018_tokyo247,Doan-2019,Hausler_2021_patch_netvlad,Zaffar-2021} and Geographic Information Systems (GIS). This typically involves how to compare images taken from the ground with photos of known locations from database that may contain a large number of landmarks, buildings, and other unique geographical elements, namely, identifying and locating geographical images taken from different perspectives \cite{jin2017learned,Liu_2019_sare,ge2020self,Ibrahimi_2021_insideout_vpr,warburg2020mapillary}. This task is crucial for numerous applications, such as urban planning, navigation systems, geographic monitoring, autonomous driving \cite{doan2019scalable}, and military reconnaissance. With the advancement of deep learning and artificial intelligence, an increasing amount of researches are dedicated to developing algorithms capable of automatically performing visual geo-localization.

Previous researches approach visual geo-localization tasks as an image retrieval problem, utilizing CNN-based models for feature extraction and similarity measurement to achieve effective localization. NetVLAD \cite{Arandjelovic_2018_netvlad} leverages convolutional neural networks to extract local features from images and aggregates these features into a global descriptor, thereby capturing the semantic information of the entire image without being affected by the positions of feature points. However, the local features extracted by these techniques may fail under extreme lighting conditions and seasonal changes, leading to significant appearance differences between query and database images. Consequently, these methods still have shortcomings in terms of feature robustness and discriminability.

\begin{figure}
\vspace {-0.5cm}
\begin{center}
\includegraphics[width=\textwidth]{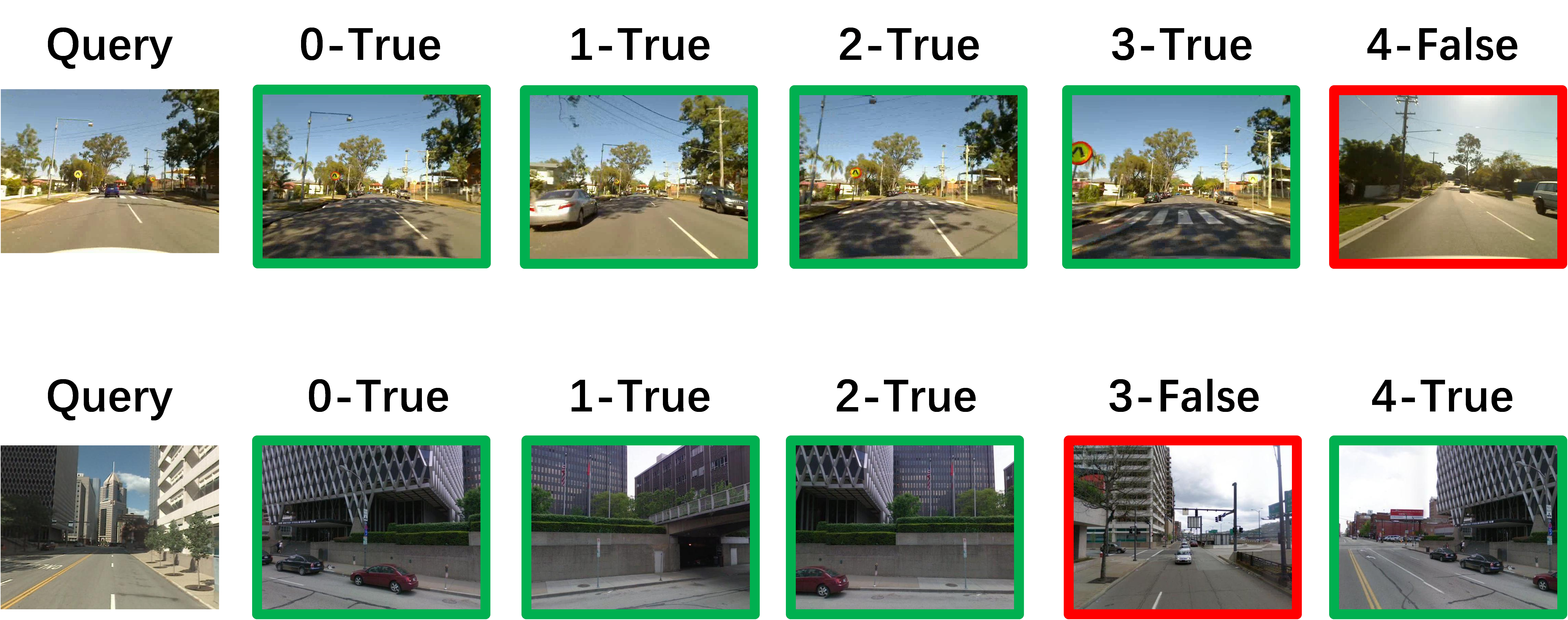}
\caption{Models employing ResNet-50 as the image encoder demonstrated results on two challenging query image datasets Pitts30k and St Lucia, revealing the top five matching results with database images.} \label{fig1}
\end{center}
\vspace{-0.8cm}
\end{figure}

Traditional methods face challenges in dealing with image content variations caused by differences in perspective, scale, and environmental changes. However, the CLIP \cite{radford2021learning} model can more effectively bridge these differences with its strong multi-modal understanding capability. However, in most visual geo-localization tasks, the lack of specific vocabulary to describe images results in the neglect of rich semantic features within the images. Consequently, the approaches involving visual-language models do not yet be widely adopted in visual geo-localization tasks. Nonetheless, the prospects for this task remain very broad. In this paper, we propose a two-stage training multi-modal visual geo-localization method named ProGEO. We plan to fully leverage the potential of CLIP to optimize for specific regions or similar types of geographic features in visual geo-localization tasks. This includes supplementing missing text descriptions, thereby enhancing the accuracy and efficiency of localization, and better generalizing this method in real-world scenarios. We use an image encoder for Resnet-50 to visualize on the Pitts30k and St Lucia datasets, as shown in Fig.~\ref{fig1}. 

In summary, our main contributions are as follows:
\begin{itemize}
\item[$\bullet$] We propose a two-stage training model ProGEO, which leverages the multi-modal capability of CLIP for visual geo-localization. By providing vague text descriptions, it addresses the issue of geographical images lacking precise descriptions which makes it difficult to widely use multi-modal models.
\item[$\bullet$] We combine representational learning and metric learning to model the details of geographic images, the introduction of which bring significant improvements to the task of visual geo-localization. Furthermore, we explore the relationship between the amount of training and accuracy by freezing some layers of the model.
\item[$\bullet$] Through a broad range of experiments, we demonstrate that our model achieves superior generalization across different datasets. It achieves competitive results on the majority of visual geo-localization datasets, validating the effectiveness and applicability of our approach. 
\end{itemize}

\section{Related Works}
\subsection{Visual Geo-localization}
Visual Geo-localization (VG) is a technology that identifies geographic locations based on image content. By analyzing visual elements in photos such as landmarks, architectural styles, natural landscapes, and even weather and lighting conditions, it matches identified features with those of known locations stored in database to determine the shooting location of the photo. Early works often use computer vision techniques to identify local features in images like repetitive structures, where algorithms like SIFT \cite{Lowe_2004_sift} and SURF \cite{Bay_2008_surf} are used to extract and compare local features of images. NetVLAD \cite{Arandjelovic_2018_netvlad} abandons the hand-crafted descriptor used in SIFT, implementing dimensionality reduction of local features through clustering and a differentiable indicator function. Patch-NetVLAD \cite{Hausler_2021_patch_netvlad} and GM-NetVLAD \cite{cao2020image} combine the advantages of local and global features to demonstrate that the integrated features exhibit high invariance to both conditions and viewpoint changes.


Unlike the NetVLAD method, GeM \cite{Radenovic_2019_gem} represents a learnable form of generalized global pooling to highlight important information in feature maps. Inspired by facial recognition, CosPlace \cite{Berton_2022_cosPlace} leverages the availability of dense data for training. It then employs the concept of classification to perform retrieval on the model. MixVPR \cite{Alibey_2023_mixvpr} integrates overall global features, utilizing feature maps from pretrained networks to capture the holistic semantic information of images. AnyLoc \cite{keetha2023anyloc} leverages off-the-shelf self-supervised models for feature extraction, combining it with unsupervised feature aggregation to obtain the sharpest and most unique feature representations. TransVPR \cite{wang2022transvpr} leverages Transformers to automatically find features in images, using self-attention mechanisms to extract image features at different semantic levels. L2LTR \cite{yang2021cross} leverages a CNN-transformer approach to facilitate the understanding and correspondence of geometric shapes between image pairs, employing inter-layer Transformer mechanisms to learn perspective transformations and semantic relationships between images. TransGeo \cite{zhu2022transgeo} employs a dual-transformer branch to directly learn effective geographical representations from original images. It leverages an attention-guided non-uniform cropping strategy to remove a significant amount of non-informative patches from database images. $R^{2}$Former \cite{zhu2023r2former} takes into account feature correlations, attention values, and $xy$ coordinates, and learns to determine whether image pairs come from the same location using the reordering module.

\subsection{Visual Language Model}
The Vision Transformer \cite{Dosovitskiy_2021_vit} (ViT) is an innovative image processing model inspired by the Transformer \cite{Vaswani_2017_attentionisallyouneed} architecture from the field of NLP (Natural Language Processing). Unlike traditional methods based on CNN, ViT divides images into a series of small patches and processes these image blocks similar to text sequences. This approach enables ViT to capture global features in images, not just local features. Visual language pretraining models are a multi-modal learning approach designed to understand and process data that includes both visual and linguistic information. These models are typically pretrained on large datasets to learn the deep associations between vision and text. After the pretraining stage, these models can be used for various downstream tasks such as Visual Question Answering \cite{cascante2022simvqa,ding2022mukea,song2022clip}(VQA) and image production \cite{vinker2022clipasso}.

The model CLIP \cite{radford2021learning} is developed by OpenAI, which is capable of recognizing new categories and concepts without specific task training. The architecture of the CLIP model consists of two main components: an image encoder and a text encoder. CoOp \cite{zhou2022learning} builds on CLIP by utilizing learnable vectors to model the contextual words in prompts. CoCoOp \cite{zhou2022learning} trains a lightweight network Meta-Net to learn the features of input images, thereby imposing constraints on the image encoder. CLIP-ReID \cite{li2023clip} utilizes a prompt tuning method similar to CoOp, leveraging the implicit correspondence between texts and images to aid in the task of person ReID. IM2City \cite{wu2022im2city} and DenseCLIP \cite{rao2022denseclip} guide dense prediction models for improved geolocation tasks by leveraging the power of pretrained language-guided fine-tuning paradigms and zero-shot learning capability. MaPLe \cite{khattak2023maple} utilizes separately learned prompts to gradually model stage feature relations, thereby generalizing to new classes, new target datasets, and new domain transfers. CPT \cite{yao2021cpt} reconstructs the visual positioning problem in images and texts using color-based co-referential markers to make it a fill-in-the-blank problem.

Our paper is heavily influenced by these multi-modal visual language model studies and extends their concepts to downstream tasks involving visual geo-localization, extracting cues about geographic locations from text information to effectively bridge the internal representations of the visual and language worlds.

\section{Method}
\begin{figure}
\vspace{-0.1cm} 
\begin{center}
\includegraphics[width=\textwidth]{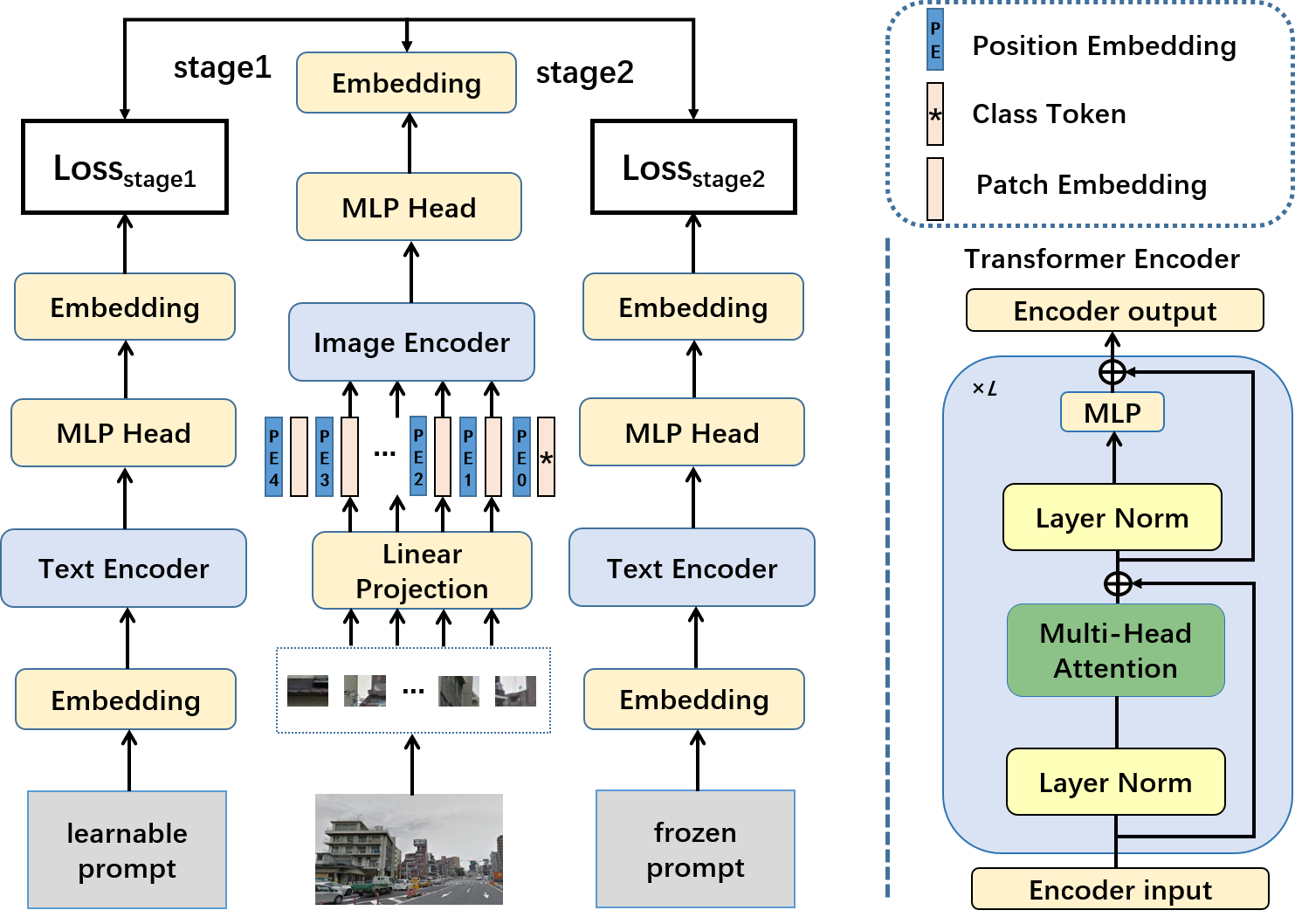}
\caption{The overall architecture of our model with a ViT image encoder backbone.} \label{fig2}
\end{center}
\vspace{-1cm} 
\end{figure}

ProGEO is a multi-modal visual geo-localization system. We first introduce the complete two-stage training process of the model. Furthermore, we elaborate on the introduced triplet loss \cite{hermans2017defense} method, providing a challenging sample mining approach to learn more reliable features. Given that CLIP comprises image encoder based on Resnet \cite{He_2016_resnet} and ViT \cite{Dosovitskiy_2021_vit}, our proposed approach is validated on both ResNet-50, Resnet-101, ViT-B/16, and ViT-B/32. As is shown in Fig.~\ref{fig2},  we introduce the overall architecture of our model with a ViT image encoder backbone, which includes a two-stage training process and the loss functions. 

\subsubsection{The First Training Stage}
\begin{figure}
\vspace{-0.3cm} 
\begin{center}
\includegraphics[width=\textwidth]{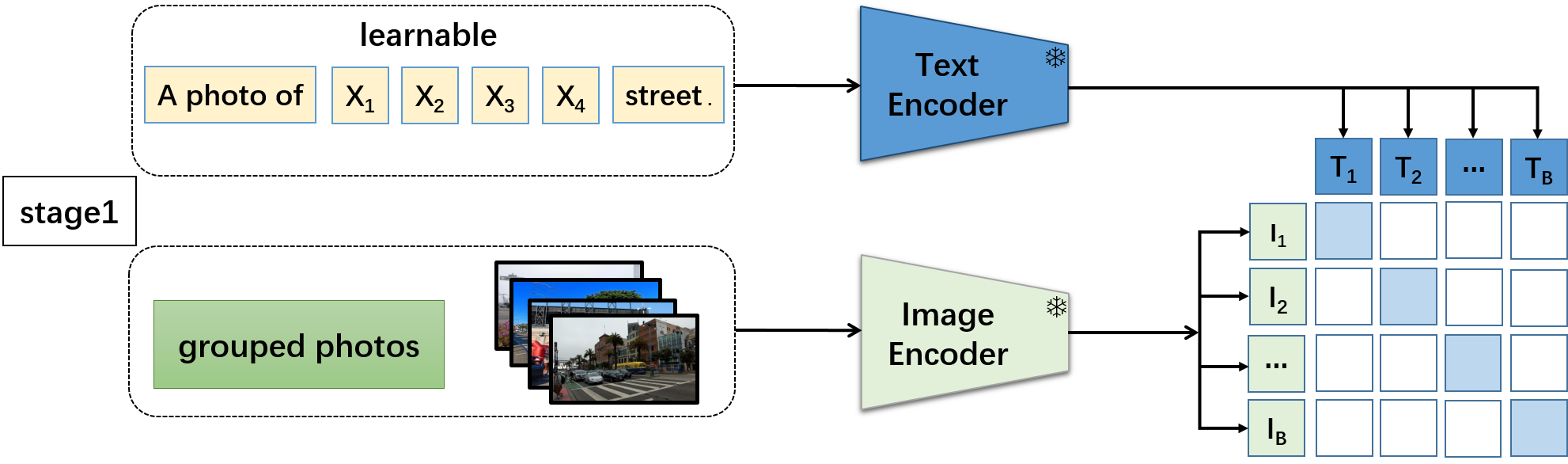}
\caption{The first training stage for model ProGEO.} \label{fig3}
\end{center}
\vspace{-0.4cm} 
\end{figure}

In the first training stage, we use learnable prompts to mine the stored hidden states of pre-trained image encoder and text encoder, allowing CLIP to retain its advantages (see Fig.~\ref{fig3}). We reference the $L_{it}$ and $L_{ti}$ in the CLIP model. Images and texts are processed through their respective encoders, using the CLS tokens of images and the EOS tokens of texts as the final feature encodings for images and texts. $i\in ({1\cdots B})$ denotes an index of images within a batch and similarity is calculated using the dot product for contrastive learning, where the diagonal represents the pairing of images and texts. Our goal is to make the similarity matrix diagonal tend towards 1, thereby making the text and image embeddings more similar.

\noindent The image-to-text contrast loss $L_{it}$ is:
\begin{equation}\label{eq1}
L_{it}(i)=-log\frac{exp(V_{i}\cdot T_{i}/\tau )}{\sum_{j=1}^{B}exp(V_{i}\cdot T_{j}/\tau )}
\end{equation}

\noindent The text-to-image contrast loss $L_{ti}$ is:
\begin{equation}\label{eq2}
L_{ti}(y_i)=\frac{-1}{|P(y_i)|}\sum_{p\subseteq P(y_i)}^{}log\frac{exp(V_{p}\cdot T_{y_i}/\tau )}{\sum_{j=1}^{B}exp(V_{j}\cdot T_{y_i}/\tau )}
\end{equation}

We first train a learnable and fuzzy text description, designing the description in the format of ``A photo of a X X X X street.'', where X represents a placeholder. In this stage, only the X in the text prompts is optimized, while the image encoder $I(\cdot)$ and text encoder $T(\cdot)$ are frozen. We input the entire train set into $I(\cdot)$, thus obtaining all image features and corresponding label information at once, preparing for the next stage of training and improving training efficiency. Assuming we have a batch of $N$ visual features $V$ and text features $T$, where 
$V_{i}$ and $T_{i}$ respectively denote the visual and text features of the i-th sample. $\cdot $ represents the dot product of a vector and the temperature parameter $\tau $ is used to adjust the sensitivity of the softmax function. By adjusting the value of $\tau $, the balance between the relative contributions of positive and negative examples during the learning process can be modulated, thereby affecting the representational capacity of the model and generalization performance. In the CLIP model, only the own transform of the sample is considered positive for a sample in a batch, while all other samples in the batch are considered negative and assuming no two identical items have the same id. In the context of visual geo-localization tasks, a batch may contain multiple image samples with the same ID and there may exist images of the same category as the sample. So we make corresponding modifications to the $L_{ti}$. For the text token of a given index, we calculate the cross-entropy loss for each positive sample image for all $p$ positive sample images in a batch. Then we average these loss values, thereby obtaining the average loss for positive samples in the entire batch to evaluate the performance of the model on positive samples, while the $L_{it}$ remains unchanged.

\noindent Therefore, the total loss in the first training stage is:
\begin{equation}\label{eq3}
L_{stage1}=L_{it}+L_{ti}
\end{equation}
 
Through optimizing the loss function in the first training stage, the model is trained to improve the correspondence between visual and text features, thereby enabling a more accurate understanding and representation of the relationship between images and texts.
\begin{figure}
\begin{center}
\includegraphics[width=\textwidth]{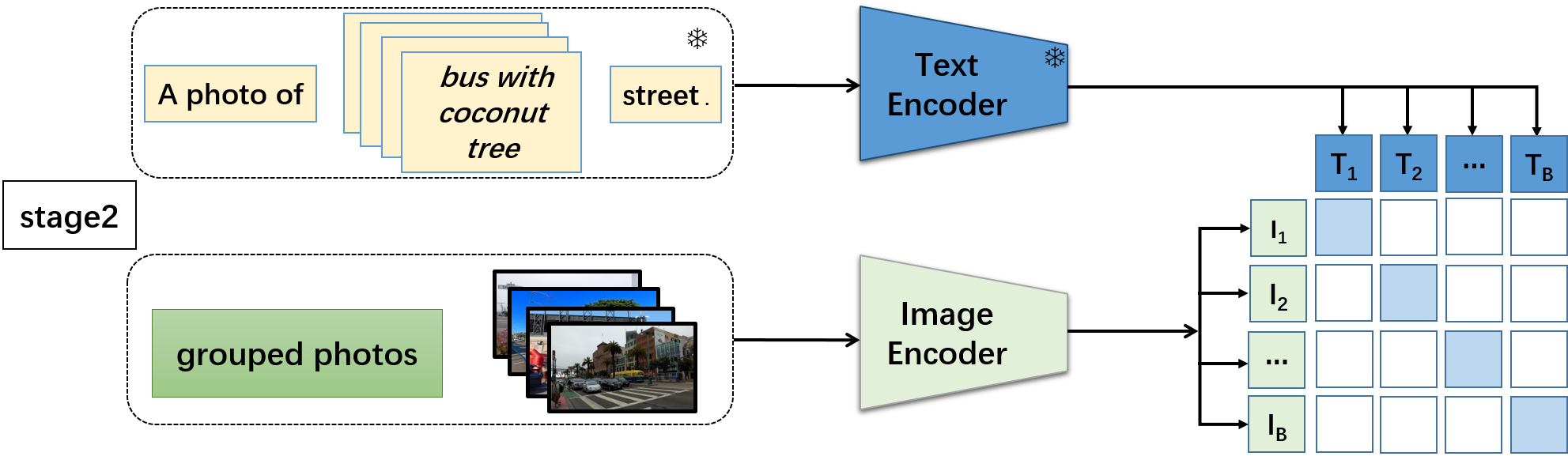}
\caption{The second training stage for model ProGEO.} \label{fig4}
\end{center}
\vspace{-1cm} 
\end{figure}
\subsubsection{The Second Training Stage}
In the second training stage, we employ the CosPlace\cite{Berton_2022_cosPlace} method to categorize the dataset using UTM coordinates $\left \{ east ,north\right \}$ to divide the database into square geographic cells. The text encoder from the first training stage remains frozen, and only the image encoder $I(\cdot)$ participates in training, converting them into semantic information of the texts. Combined with the extracted semantic features, the content of the images is controlled and guided to a certain degree (see Fig.~\ref{fig4}). $k$ is the number of categories and $q_k=(1-\epsilon)\delta_{k,y}+\epsilon/N$ denotes value in the target distribution. We utilize the text features obtained after the first training stage to calculate the image-to-text cross-entropy loss, fully leveraging the multi-modal capability of CLIP:
\begin{equation}\label{eq4}
L_{ce}(i)=\sum_{k=1}^{N} -q_k\,log\frac{e^{(V_i\cdot T_{y_k}/\tau )} }{ {\textstyle \sum_{y_{j=1}}^{N}e^{(V_i\cdot T_{y_j}/\tau )}}}
\end{equation}
\noindent The label space within the dataset is continuous in the VG task, making categorization non-intuitive. Inspired by CosPlace \cite{Berton_2022_cosPlace}, we divide the continuous label space into distinct classes and iterate over each group, training each group in sequence. We employ four hyperparameters for constructing groups, such as $M$, $\propto $, $N$ and $L$. No two connected classes belong to the same group and a category cell is explicitly assigned to $L$ groups. Formally, each group $G_{uvw}$ is defined as follows:
\begin{equation}\label{eq5}
G_{uvw}=\left \{ C_{{e_i}{n_j}{h_k}}: (e_i\,mod\,N=u)\wedge (n_j\,mod\,N=v)\wedge(h_k\,mod\,L=w)\right \} 
\end{equation}
\noindent By partitioning the above datasets, each group is treated as a separate dataset. Each group performs the Large
Margin Cosine Loss \cite{Wang_2018_cosFace} (LMCL) in the face recognition technology CosFace \cite{Wang_2018_cosFace}, and we train each group in turn:

\begin{equation}\label{eq6}
L_{cos}=L_{lmcl}(G_{uvw})
\end{equation}

On the basis of the representation learning of CosFace \cite{Wang_2018_cosFace}, this retrieval problem can also be addressed using standard metric learning. The triplet loss \cite{hermans2017defense} function is a commonly used method to learn embeddings in a space, measuring the similarity between two samples using functions that compute Euclidean and cosine distances. We validate its effectiveness in experiments, where the triplet loss is defined as follows:
\begin{equation}\label{eq7}
L_{triplet}=max(d(a,p)-d(a,n)+margin,0)
\end{equation}

\noindent The input is a triplet, including an anchor example, a positive  example, and a negative example. By optimizing the process such that the distance between the anchor example and the positive example is smaller than the distance between the anchor example and the negative example, the calculation of similarity between samples is realized. $d(a,p)$ and $d(a,n)$ respectively represent the distance between feature embedding of the anchor point $a$ and the positive example $p$ and the negative example $n$. $margin$ is a constant greater than 0, the final optimization objective is to reduce the distance between the features of the target image and the positive example, and to increase the distance between the features of the target image and the negative example.
Therefore, our total loss for the second training stage is:
\begin{equation}\label{eq8}
L_{stage2}=L_{ce}+L_{cos}+L_{triplet}
\end{equation}

\section{Experiments}
In section \ref{section4.1}, we provide the details of the datasets and the evaluation metrics used in the experiments. In section \ref{section4.2}, we provide other relevant implementation details. In section \ref{section4.3}, we demonstrate the test results of our model on multi-view datasets and frontal-view datasets. In section \ref{section4.4}, we conduct ablation studies to verify the effectiveness of our model. 
\subsection{Datasets and Evaluation Metrics}\label{section4.1}
\subsubsection{Datasets}
We use 7 datasets to evaluate our method, which together cover a variety of real-world scenes. Pitts30k \cite{Arandjelovic_2018_netvlad}, Pitts250k \cite{Arandjelovic_2018_netvlad}, Tokyo24/7 \cite{Torii_2018_tokyo247}, SF-XL test v1 \cite{Berton_2022_cosPlace} and SF-XL test v2 \cite{Berton_2022_cosPlace} belong to multi-view datasets, where the query images cover sidewalk pictures collected with mobile phones, and the database images come from street view images. Meanwhile, St Lucia \cite{Milford_2008_st_lucia} and Mapillary Street Level Sequences (MSLS) \cite{warburg2020mapillary} belong to frontal-view datasets, containing the majority of images along roadsides. Our model is trained on the dataset SF-XL \cite{Berton_2022_cosPlace} (San Francisco eXtra Large) which is a city-wide, dense, and temporally varied dataset labeled with GPS coordinates and headings. SF-XL has a total of 41.2 million images which is created from Google Street View images, covering the entire area of San Francisco. These photos were taken between 2009 and 2021 which provides a large amount of long-term temporal variation, we use these 41.2 million images as a train set. To save time and GPU resources during testing, we only use 2.8M images as the test set database and we use two different sets of test queries SF-XL test v1 and SF-XL test v2.

\subsubsection{Evaluation Metrics}
Following common practice, we adhere to the standard geolocation recognition evaluation procedure using recall rates Rank-1 (R@1) and Rank-5 (R@5) to assess performance. The Rank-N recall rate indicates the frequency of correct matches appearing in the top $N$ results returned by the model in the test set, used to evaluate the coverage of the detector of all targets to be detected. We assume a positive threshold distance of 25 meters, within which range the query image is considered to be correctly located.
\subsection{Implementation Details}\label{section4.2}
\subsubsection{Image Backbone}
Our model ProGEO employs the image encoder $I(\cdot)$ of CLIP as the backbone network. The image encoders are Resnet-50, Resnet-101, ViT-B/16, and ViT-B/32, while the text encoder utilizes BERT \cite{lee2018pre}. ViT-B/16 and ViT-B/32 feature 12 transformer layers, with each multi-head attention block comprising 6 heads, which are initialized with non-pretrained weights on ImageNet-1K. For images of different resolutions, the positional encoding is expanded through interpolation methods to cover more positional information when fine-tuning the ViT. The feature dimensions of these models are consistent with CLIP, so the output dimension of Resnet-50 is set to 1024. For Resnet-101, ViT-B/16 and ViT-B/32, it is set to 512. During the first training stage, the input image size is adjusted to 224$\times$224, consistent with the resolution of the pretrained CLIP model. In the second training stage, the image size is set to 512$\times$512. Our approach is implemented in the PyTorch deep learning framework and trained on a single device equipped with a 4090 model GPU and 24GB of VRAM.
\subsubsection{Training Details}
In our experiments with the mentioned datasets, we pass only the learnable parameters of unfrozen layers to the Adam optimizer \cite{Kingma_2014_adam}, with an initial learning rate of 0.01 in the first training stage, incorporating official cosine annealing decay of PyTorch. The first training stage focuses on optimizing the learnable text prompts X using contrastive loss, conducting training over 480 epochs with a batchsize of 512. For the second training stage, the training of the image encoder is supported by the text encoder from the first training stage, with a learning rate of 0.0001 and a classification head learning rate of 0.01. Hyperparameters are established as $M=10$, $\propto =60^{\circ}$, $N=3$ and $L=2$ resulting in 50 groups, each containing approximately 35k classes, with an average of 19.8 images per class. Each epoch involves iterating through one group 10k times with one group per epoch, for a total of 64 epochs trained at a batchsize of 32. To ensure each group is revisited multiple times during training, we extract 8 sets of image features and corresponding label information at once. The sets from 0 to 7 undergo cyclic training, then the image feature of the current set is obtained to calculate the text features for each variation. The process involves validating once after each training iteration. In the final inference phase, images only need to pass through the trained image encoder to obtain image embeddings, eliminating the need for using the text encoder.

\subsection{Comparison with State-of-the-Art Methods}\label{section4.3}

We conduct a series of extensive experiments to assess the reliability of our method and compare our model with state-of-the-art models. To appropriately evaluate the results, we test our approach on 7 datasets: Pitts250k, Pitts30k, Tokyo24/7, St Lucia, MSLS, SF-XL test v1, and SF-XL test v2. For the MSLS dataset, considering that the test set labels do not be publicly released, we conduct our tests on the validation set. To ensure fair comparisons, we utilize recall rates R@1 and R@5 with a 25-meter threshold as the measurement metric. The image encoder of our CNN model is Resnet-101 and the image encoder of our Transformer model is ViT/B-16. The results of our visual geo-localization test sets for various models are presented in Table~\ref{table:headings}, we significantly surpass many of the current single-modal visual geo-localization methods. Our method is more computationally efficient and the detailed comparisons will be provided in the following section.

\begin{table}
\centering
\vspace{-0.3cm} 
\caption{The R@1 and R@5 on multi-view datasets and frontal-view datasets are segmented according to the train sets used and the feature dimensions, the threshold distance for positivity is 25 meters. The best results for each dataset are shown in bold.}
\label{table:headings}
\begin{adjustbox}{width=\textwidth}
\footnotesize
\begin{tabular}{lccccccccccccccccccccccccc}
\toprule
\multicolumn{1}{l}{\multirow{2}{*}{Method}} & \multicolumn{1}{c}{\multirow{2}{*}{Desc. dim.}} & \multicolumn{1}{c}{\multirow{2}{*}{Train set}}  & \multicolumn{2}{c}{Pitts30k} & & \multicolumn{2}{c}{Pitts250k} && \multicolumn{2}{c}{Tokyo 24/7} & & \multicolumn{2}{c}{MSLS} & & \multicolumn{2}{c}{St Lucia} & & \multicolumn{2}{c}{Average} \\
\cline{4-5} \cline{7-8} \cline{10-11} \cline{13-14} \cline{16-17} \cline{19-20}
\multicolumn{3}{c}{}
& R@1  & R@5  & & R@1  & R@5  & & R@1  & R@5 &
& R@1  & R@5  & & R@1  & R@5  & & R@1  & R@5 \\
\hline
NetVLAD\cite{Arandjelovic_2018_netvlad} & 32768 & Pitts30k
& 86.1 & 94.1 && 85.9 & 93.6 && 62.2 & 75.4 && 54.8 & 66.6 && 70.8 & 81.8 & & 72.0 & 82.3 \\
NetVLAD\cite{Arandjelovic_2018_netvlad} & 32768 & MSLS
& 80.9 & 90.6 && 79.7 & 90.2 && 63.6 & 77.5 && 75.4 & 84.2 && 92.8 & 97.6 & & 78.5 & 88.0 \\
CRN\cite{jin2017learned}  & 32768 & Pitts30k
& 86.3 & 94.6 && 87.0 & 94.5 && 62.8 & 77.4 && 57.6 & 70.4 && 70.9 & 82.8 & & 72.9 & 83.9 \\
SPE-VLAD\cite{Yu_2020_SPEVlad}  & 32768 & Pitts30k
&  -   &  89.2 & &  -   & - & &  -   & 63.9 &
&  -   &  -    & &  -   &  -   & & - &  -  \\
SARE\cite{Liu_2019_sare} &  4096 & Pitts30k
& 87.2 & 93.9 && 88.0 & 94.8 && 74.8 & 84.3 && 62.4 & 73.2 && 72.7 & 86.0 & & 77.0 & 86.4 \\
SFRS\cite{ge2020self}  &  4096 & Pitts30k
& 88.7 & 94.2 && 90.1 & 95.8 && 78.5 & 87.3 && 62.8 & 73.0 && 72.5 & 85.4 & & 78.5 & 87.1 \\
SRALNet\cite{Peng_2021_sralNet} & 4096 & Pitts30k
&  -   &  -   && 87.8 & 94.8 && 72.1 & 83.2 &
&  -   &  -   &&  -   &  -   &&  -   &  -   \\
APPSVR\cite{Peng_2021_appsvr} & 4096 & Pitts30k
& 87.4 & 94.3 && 88.8 & 95.6 && 77.1 & 85.7 &
&  -   &  -   &&  -   &  -   &&  -   &  -   \\
GeM\cite{Radenovic_2019_gem}  &   512 & Pitts30k & 
77.9 & 90.5 && 75.3 & 88.4 && 46.4 & 65.3 && 51.8 & 64.4 && 59.9 & 76.3 & & 62.3 & 77.0 \\
GeM\cite{Radenovic_2019_gem}  & 512 & MSLS
& 71.6 & 85.1 && 65.3 & 81.0 && 44.9 & 62.6 && 66.7 & 78.9 && 84.6 & 93.3 & & 66.6 & 80.2 \\
APANet\cite{Zhu_2018_apanet}  & 512 & Pitts30k
&  -   &  -   && 83.7 & 92.6 && 67.0 & 81.0 &
&  -   &  -   & &  -  &  -  & &  -   &  -   \\
CosPlace\cite{Berton_2022_cosPlace} & 512 & SF-XL  
& 88.5 & 94.5 && 89.7 & 96.4 && 82.8 & 90.0 && 79.5 & 87.2 && 94.3 & 97.4 && 87.0 & 93.1 \\
Conv-AP\cite{Alibey_2022_gsvcities} & 512 & GSV-Cities 
& 89.1 & - && 90.4 & - && 61.3 & - &&
83.6 & - && 99.2 &- && 84.7 & - \\
MixVPR\cite{Alibey_2023_mixvpr} & 512 & GSV-Cities 
& 90.4 & - && \textbf{93.0} & - && 78.4 & - && 83.6 & - && 99.2 & - && 88.9 & - \\
$R^{2} $Former\cite{zhu2023r2former} & 256 & MSLS & 91.1 & 95.2 && - & - && \textbf{88.6} & 91.4 && 73.0 & 85.9 && \textbf{99.7} & - && 88.1 & 90.8 \\
\hline
\textbf{ProGeo(CNN)$^{*}$} &  512 &  SF-XL 
& 91.8 & 97.4 && 92.2 & \textbf{97.7} && 87.0 & 92.7 && 84.8 & 91.4 && \textbf{99.7} & \textbf{99.9} && 91.1 & \textbf{96.0}\\
\textbf{ProGeo(Transformer)$^{*}$} & 512 & SF-XL 
& \textbf{93.0} & \textbf{98.3} && 90.7 & 95.9 && \textbf{88.6} & \textbf{93.3} && \textbf{84.9} & \textbf{91.6} && 99.5 & \textbf{99.9} && \textbf{91.3} & 95.8\\
\bottomrule
\end{tabular}
\end{adjustbox}
\vspace{-0.5cm}
\end{table}

\subsection{Ablation Study}\label{section4.4}
To explore the impact of different modules on the overall performance of the matching task, we conduct ablation studies on the VG datasets. We use a method that freezes half of the image encoder and does not add prompts or triplet loss as the baseline. We gradually integrate into our approach from this starting point, assessing based on key indicators such as R@1 and R@5.
\subsubsection{Backbones}
In our extensive experiments detailed in the tables, we investigate the impact of using different backbones on the results. Our method employs Resnet-50, Resnet-101, ViT/B-16, and ViT/B-32 as the image encoder to test on the SF-XL test set. ViT/B-16 as the image encoder performs best and the result is given in Table~\ref{table:second}.
\vspace{-0.3cm}
\begin{table}
\centering
\caption{Ablation study on different backbones. }
\label{table:second}
\footnotesize
\begin{tabular}{lcccccccccccccccccccccccc}
\hline
\multirow{2}{*}{Method} & \multirow{2}{*}{Backbone} & \multirow{2}{*}{Desc.dim}  & \multirow{2}{*}{Params} & 
\multirow{2}{*}{GFLOPS} & \multicolumn{2}{c}{SF-XL test v1} && \multicolumn{2}{c}{SF-XL test v2} \\
\cline{6-7} \cline{9-10} 
\multicolumn{3}{c}{}
&&& R@1  & R@5  && R@1  & R@5\\
\hline
CosPlace\cite{Berton_2022_cosPlace} & Vgg-16 & 512 & $58\times 10^{6}$ & 16 & 64.7 & 73.3 && 83.4 & 91.6\\
CosPlace\cite{Berton_2022_cosPlace} & Resnet-50 & 512 & $25.5 \times 10^{6}$ & 4.1 & 76.7 & - && 89.0 & -\\
\hline
ProGEO & Resnet-50 & 1024 & $25.5 \times 10^{6}$ & 4.1 & 81.7 & 87.1 && 92.6 & 95.7\\ 
ProGEO & Resnet-101 & 512 &$45.3 \times 10^{6}$ & 8 & 83.7 & 88.4 && 93.1 & 96.3\\
ProGEO & ViT-B/32 & 512 & $88 \times 10^{6}$ & 4.41 & 78.2 & 84.8 && 89.6 & 94.8\\
ProGEO & ViT-B/16 & 512 & $86 \times 10^{6}$ & 17.56 & \textbf{84.7} & \textbf{90.3} && \textbf{93.0} & \textbf{96.7} \\
\hline
\bottomrule
\end{tabular}
\label{tab:onestageupdate}
\vspace{-0.8cm}
\end{table}

\subsubsection{Prompt and Triplet}
Building on the baseline, we combine the unfrozen image encoder with the addition of learnable prompts and triplet loss methods to validate the impact of the neural network-extracted features on visual geo-localization performance. Table~\ref{table:third} illustrates the R@1 and R@5 results of the baseline and our progressively integrated methods on the SF-XL validation set. The models incorporating our method achieve more accurate retrieval results, while the baselines struggle to correctly identify the correspondences between different patterns. This success is primarily due to our gradual addition of learnable prompts and metric learning based on hard instance mining which enhance the ability to capture similarities and differences between images during the training process. This improvement is crucial for distinguishing between different geographic locations in the feature space. By directing the model to focus more on relevant aspects for the image retrieval task, we focus on ensuring tight matches for precise visual geo-localization.

As can be seen from the Table~\ref{table:third}, ProGEO simulates the details of the geographic images more accurately and enhances the ability to capture similarities and differences between images during the training process. This approach significantly improves the ability to distinguish between different geographic locations in the feature space.
\vspace{-0.3cm}
\begin{table}
\centering
\caption{Ablation on prompt, frozen or not and triplet loss.}
\footnotesize
\label{table:third}
\begin{tabular}{lccccccccccccccccccccc}
\toprule
\multicolumn{3}{c}{\multirow{2}{*}{Method}} & \multirow{2}{*}{unfrozen} & \multirow{2}{*}{triplet} & \multicolumn{2}{c}{Resnet-50} & & \multicolumn{2}{c}{Resnet-101} & & \multicolumn{2}{c}{ViT/B-16} & & \multicolumn{2}{c}{ViT/B-32}\\
\cline{6-7} \cline{9-10} \cline{12-13} \cline{15-16} 
\multicolumn{3}{c}{} & & & R@1  & R@5  & & R@1  & R@5  & & R@1  & R@5 & & R@1  & R@5 \\
\hline
\multicolumn{3}{c}{baseline}  &&& 94.0 & 97.5 && 94.3 & 98.0 && 83.9 & 88.7 && 88.6 & 95.0\\ 
\hline
\multirow{4}{*}{\centering baseline+prompt} & \multirow{4}{*}{\centering} &&&& 94.8 & 98.1 && 95.0 & 98.2 && 92.1 & 96.5 && 91.6 & 96.1\\ 
& & &\checkmark && 95.1 & 98.1 && 95.2 &98.4 && \textbf{94.1} & \textbf{97.7} && 92.5 & 96.9\\
& & & & \checkmark & 94.7 & 98.1 && 94.6 & 98.3 && 92.7& 97.0 && 91.5& 96.4\\
& & & \checkmark & \checkmark & \textbf{95.6} & \textbf{98.2} && \textbf{95.7} & \textbf{98.5} && 94.0 & 97.6 && \textbf{92.8} & \textbf{97.1}\\
\hline
\bottomrule
\end{tabular}
\vspace{-0.5cm}
\end{table}

\subsubsection{Number of Frozen Layers}
\begin{figure}
\vspace{-0.5cm} 
\begin{center}
\includegraphics[width=0.8\textwidth]{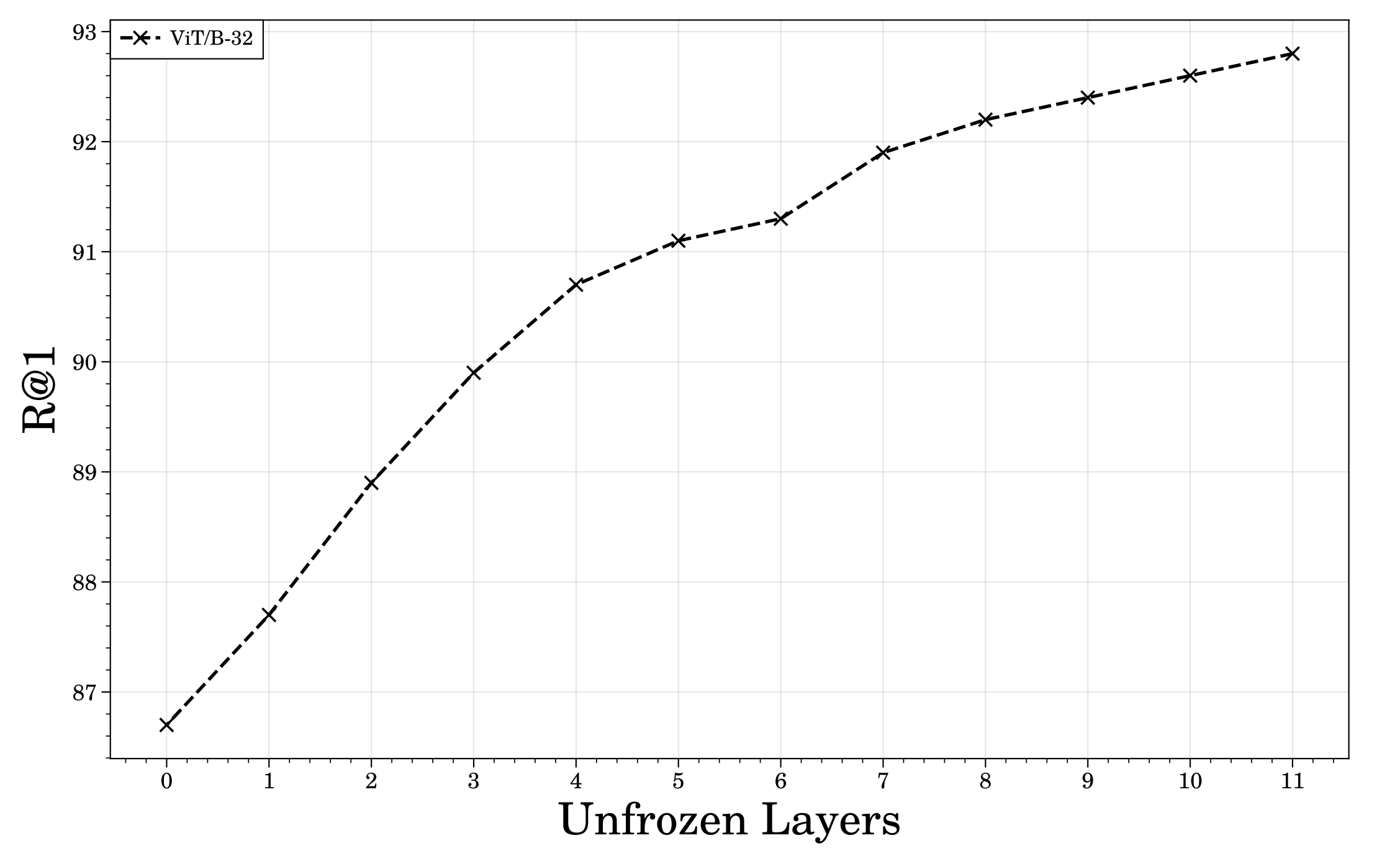}
\caption{Ablation on the number of frozen layers for ViT/B-32 image encoder.} \label{fig5}
\end{center}
\vspace{-1cm} 
\end{figure}

In the overall model framework, we explore the impact of freezing layers to prevent them from participating in training on experimental outcomes. When layers are not frozen, the parameters of the learnable parameter layers in the model change as training progresses. Although freezing layers of the model can effectively reduce the computational cost during the training process, it may lead to diminished performance. Therefore, a trade-off between accuracy and computational cost should be considered in practical applications. We freeze 0-11 layers of ViT-B/32 on the SF-XL validation set and the R@1 results are shown in Fig.~\ref{fig5}.

\section{Conclusions}
The purpose of the visual geo-localization task is to map query images to a feature space alongside database images, identifying locations from various angles. Fine-grained geographic images lack detailed text descriptions. Despite substantial progress in recent studies, performance significantly declines in real-world scenarios. To address these issues, We introduce a visual geo-localization method ProGEO, based on the vision-language model CLIP. In this system, we discover that fine-tuning a visual model initialized by the CLIP image encoder achieves excellent performance in our task. We design a two-stage training strategy to foster better visual representation. In the first training stage, we fully leverage the multi-modal descriptive capability within CLIP through a set of learnable text prompts. In the second training stage, we use the text prompts to assist in training the image encoder, establishing a linkage channel between texts and images, which further improves matching performance. Moreover, by introducing triplet loss, we endow the model with more robust performance and excellent generalizability to other domains. Despite the simplicity of the approach, ProGEO achieves remarkably good outcomes. Experiments demonstrate that our proposed strategy achieves competitive results on the majority of visual geo-localization datasets.
{\small
\bibliographystyle{splncs04}
\bibliography{mybibliography}

\begin{thebibliography}{10}
\providecommand{\url}[1]{\texttt{#1}}
\providecommand{\urlprefix}{URL }
\providecommand{\doi}[1]{https://doi.org/#1}

\bibitem{Alibey_2022_gsvcities}
Ali-bey, A., Chaib-draa, B., Gigu{\`e}re, P.: Gsv-cities: Toward appropriate supervised visual place recognition. Neurocomputing  \textbf{513},  194--203 (2022)

\bibitem{Alibey_2023_mixvpr}
Ali-bey, A., Chaib-draa, B., Gigu{\`e}re, P.: Mixvpr: Feature mixing for visual place recognition. In: Proceedings of the IEEE/CVF Winter Conference on Applications of Computer Vision. pp. 2998--3007 (2023)

\bibitem{Arandjelovic_2018_netvlad}
{Arandjelović}, R., Gronat, P., Torii, A., Pajdla, T., Sivic, J.: {NetVLAD}: {CNN} architecture for weakly supervised place recognition  \textbf{40}(6),  1437--1451 (2018). \doi{10.1109/TPAMI.2017.2711011}

\bibitem{Bay_2008_surf}
Bay, H., Ess, A., Tuytelaars, T., Van~Gool, L.: Speeded-up robust features (surf). Computer Vision and Image Understanding  \textbf{110},  346--359 (06 2008). \doi{10.1016/j.cviu.2007.09.014}

\bibitem{Berton_2022_cosPlace}
Berton, G., Masone, C., Caputo, B.: Rethinking visual geo-localization for large-scale applications. In: CVPR (June 2022)

\bibitem{cao2020image}
Cao, Y., Zhang, J., Yu, J.: Image retrieval via gated multiscale netvlad for social media applications. IEEE MultiMedia  \textbf{27}(4),  69--78 (2020)

\bibitem{cascante2022simvqa}
Cascante-Bonilla, P., Wu, H., Wang, L., Feris, R.S., Ordonez, V.: Simvqa: Exploring simulated environments for visual question answering. In: Proceedings of the IEEE/CVF Conference on Computer Vision and Pattern Recognition. pp. 5056--5066 (2022)

\bibitem{ding2022mukea}
Ding, Y., Yu, J., Liu, B., Hu, Y., Cui, M., Wu, Q.: Mukea: Multimodal knowledge extraction and accumulation for knowledge-based visual question answering. In: Proceedings of the IEEE/CVF Conference on Computer Vision and Pattern Recognition. pp. 5089--5098 (2022)

\bibitem{Doan-2019}
Doan, A.D., Latif, Y., Chin, T.J., Liu, Y., Do, T.T., Reid, I.: Scalable place recognition under appearance change for autonomous driving. pp. 9319--9328 (October 2019)

\bibitem{doan2019scalable}
Doan, A.D., Latif, Y., Chin, T.J., Liu, Y., Do, T.T., Reid, I.: Scalable place recognition under appearance change for autonomous driving. In: Proceedings of the IEEE/CVF International Conference on Computer Vision. pp. 9319--9328 (2019)

\bibitem{Dosovitskiy_2021_vit}
Dosovitskiy, A., Beyer, L., Kolesnikov, A., Weissenborn, D., Zhai, X., Unterthiner, T., Dehghani, M., Minderer, M., Heigold, G., Gelly, S., Uszkoreit, J., Houlsby, N.: {An Image is Worth 16x16 Words: Transformers for Image Recognition at Scale}. ArXiv  \textbf{abs/2010.11929} (2021)

\bibitem{ge2020self}
Ge, Y., Wang, H., Zhu, F., Zhao, R., Li, H.: Self-supervising fine-grained region similarities for large-scale image localization. In: Computer Vision--ECCV 2020: 16th European Conference, Glasgow, UK, August 23--28, 2020, Proceedings, Part IV 16. pp. 369--386. Springer (2020)

\bibitem{Hausler_2021_patch_netvlad}
Hausler, S., Garg, S., Xu, M., Milford, M., Fischer, T.: Patch-netvlad: Multi-scale fusion of locally-global descriptors for place recognition. pp. 14141--14152 (2021)

\bibitem{He_2016_resnet}
{He}, K., {Zhang}, X., {Ren}, S., {Sun}, J.: Deep residual learning for image recognition. pp. 770--778 (2016). \doi{10.1109/CVPR.2016.90}

\bibitem{hermans2017defense}
Hermans, A., Beyer, L., Leibe, B.: In defense of the triplet loss for person re-identification. arXiv preprint arXiv:1703.07737  (2017)

\bibitem{Ibrahimi_2021_insideout_vpr}
Ibrahimi, S., van Noord, N., Alpherts, T., Worring, M.: Inside out visual place recognition (2021)

\bibitem{jin2017learned}
Jin~Kim, H., Dunn, E., Frahm, J.M.: Learned contextual feature reweighting for image geo-localization. In: Proceedings of the IEEE Conference on Computer Vision and Pattern Recognition. pp. 2136--2145 (2017)

\bibitem{keetha2023anyloc}
Keetha, N., Mishra, A., Karhade, J., Jatavallabhula, K.M., Scherer, S., Krishna, M., Garg, S.: Anyloc: Towards universal visual place recognition. IEEE Robotics and Automation Letters  (2023)

\bibitem{khattak2023maple}
Khattak, M.U., Rasheed, H., Maaz, M., Khan, S., Khan, F.S.: Maple: Multi-modal prompt learning. In: Proceedings of the IEEE/CVF Conference on Computer Vision and Pattern Recognition. pp. 19113--19122 (2023)

\bibitem{Kingma_2014_adam}
Kingma, D., Ba, J.: Adam: A method for stochastic optimization  (12 2014)

\bibitem{lee2018pre}
Lee, J., Toutanova, K.: Pre-training of deep bidirectional transformers for language understanding. arXiv preprint arXiv:1810.04805  \textbf{3}, ~8 (2018)

\bibitem{li2023clip}
Li, S., Sun, L., Li, Q.: Clip-reid: exploiting vision-language model for image re-identification without concrete text labels. In: Proceedings of the AAAI Conference on Artificial Intelligence. vol.~37, pp. 1405--1413 (2023)

\bibitem{Liu_2019_sare}
Liu, L., Li, H., Dai, Y.: {Stochastic Attraction-Repulsion Embedding for Large Scale Image Localization} (2019)

\bibitem{Lowe_2004_sift}
Lowe, D.G.: Distinctive image features from scale-invariant keypoints. Int. J. Comput. Vision  \textbf{60}(2),  91--110 (2004), \url{http://dx.doi.org/10.1023/B:VISI.0000029664.99615.94}

\bibitem{Milford_2008_st_lucia}
Milford, M., Wyeth, G.: Mapping a suburb with a single camera using a biologically inspired slam system  \textbf{24},  1038--1053 (2008)

\bibitem{Peng_2021_sralNet}
Peng, G., Yue, Y., Zhang, J., Wu, Z., Tang, X., Wang, D.: Semantic reinforced attention learning for visual place recognition. pp. 13415--13422. {IEEE} (2021)

\bibitem{Peng_2021_appsvr}
Peng, G., Zhang, J., Li, H., Wang, D.: Attentional pyramid pooling of salient visual residuals for place recognition. pp. 885--894 (October 2021)

\bibitem{Radenovic_2019_gem}
Radenovi{\'c}, F., Tolias, G., Chum, O.: {Fine-tuning {CNN} Image Retrieval with No Human Annotation}  (2018)

\bibitem{radford2021learning}
Radford, A., Kim, J.W., Hallacy, C., Ramesh, A., Goh, G., Agarwal, S., Sastry, G., Askell, A., Mishkin, P., Clark, J., et~al.: Learning transferable visual models from natural language supervision. In: International conference on machine learning. pp. 8748--8763. PMLR (2021)

\bibitem{rao2022denseclip}
Rao, Y., Zhao, W., Chen, G., Tang, Y., Zhu, Z., Huang, G., Zhou, J., Lu, J.: Denseclip: Language-guided dense prediction with context-aware prompting. In: Proceedings of the IEEE/CVF Conference on Computer Vision and Pattern Recognition. pp. 18082--18091 (2022)

\bibitem{song2022clip}
Song, H., Dong, L., Zhang, W.N., Liu, T., Wei, F.: Clip models are few-shot learners: Empirical studies on vqa and visual entailment. arXiv preprint arXiv:2203.07190  (2022)

\bibitem{Torii_2018_tokyo247}
{Torii}, A., {Arandjelović}, R., {Sivic}, J., {Okutomi}, M., {Pajdla}, T.: 24/7 place recognition by view synthesis  \textbf{40}(2),  257--271 (2018)

\bibitem{Vaswani_2017_attentionisallyouneed}
Vaswani, A., Shazeer, N., Parmar, N., Uszkoreit, J., Jones, L., Gomez, A.N., Kaiser, L., Polosukhin, I.: Attention is all you need. p. 6000–6010. NIPS'17 (2017)

\bibitem{vinker2022clipasso}
Vinker, Y., Pajouheshgar, E., Bo, J.Y., Bachmann, R.C., Bermano, A.H., Cohen-Or, D., Zamir, A., Shamir, A.: Clipasso: Semantically-aware object sketching. ACM Transactions on Graphics (TOG)  \textbf{41}(4),  1--11 (2022)

\bibitem{Wang_2018_cosFace}
Wang, H., Wang, Y., Zhou, Z., Ji, X., Gong, D., Zhou, J., Li, Z., Liu, W.: Cosface: Large margin cosine loss for deep face recognition. pp. 5265--5274. Computer Vision Foundation / {IEEE} Computer Society (2018)

\bibitem{wang2022transvpr}
Wang, R., Shen, Y., Zuo, W., Zhou, S., Zheng, N.: Transvpr: Transformer-based place recognition with multi-level attention aggregation. In: Proceedings of the IEEE/CVF Conference on Computer Vision and Pattern Recognition. pp. 13648--13657 (2022)

\bibitem{warburg2020mapillary}
Warburg, F., Hauberg, S., Lopez-Antequera, M., Gargallo, P., Kuang, Y., Civera, J.: Mapillary street-level sequences: A dataset for lifelong place recognition. In: Proceedings of the IEEE/CVF conference on computer vision and pattern recognition. pp. 2626--2635 (2020)

\bibitem{wu2022im2city}
Wu, M., Huang, Q.: Im2city: image geo-localization via multi-modal learning. In: Proceedings of the 5th ACM SIGSPATIAL International Workshop on AI for Geographic Knowledge Discovery. pp. 50--61 (2022)

\bibitem{yang2021cross}
Yang, H., Lu, X., Zhu, Y.: Cross-view geo-localization with layer-to-layer transformer. Advances in Neural Information Processing Systems  \textbf{34},  29009--29020 (2021)

\bibitem{yao2021cpt}
Yao, Y., Zhang, A., Zhang, Z., Liu, Z., Chua, T.S., Sun, M.: Cpt: Colorful prompt tuning for pre-trained vision-language models. arXiv preprint arXiv:2109.11797  (2021)

\bibitem{Yu_2020_SPEVlad}
Yu, J., Zhu, C., Zhang, J., Huang, Q., Tao, D.: Spatial pyramid-enhanced netvlad with weighted triplet loss for place recognition. IEEE Transactions on Neural Networks and Learning Systems  \textbf{31}(2),  661--674 (2020). \doi{10.1109/TNNLS.2019.2908982}

\bibitem{Zaffar-2021}
Zaffar, M., Garg, S., Milford, M., Kooij, J., Flynn, D., McDonald-Maier, K., Ehsan, S.: {VPR-Bench}: An open-source visual place recognition evaluation framework with quantifiable viewpoint and appearance change  \textbf{129}(7),  2136--2174 (2021)

\bibitem{zhou2022learning}
Zhou, K., Yang, J., Loy, C.C., Liu, Z.: Learning to prompt for vision-language models. International Journal of Computer Vision  \textbf{130}(9),  2337--2348 (2022)

\bibitem{zhu2022transgeo}
Zhu, S., Shah, M., Chen, C.: Transgeo: Transformer is all you need for cross-view image geo-localization. In: Proceedings of the IEEE/CVF Conference on Computer Vision and Pattern Recognition. pp. 1162--1171 (2022)

\bibitem{zhu2023r2former}
Zhu, S., Yang, L., Chen, C., Shah, M., Shen, X., Wang, H.: R2former: Unified retrieval and reranking transformer for place recognition. In: Proceedings of the IEEE/CVF Conference on Computer Vision and Pattern Recognition. pp. 19370--19380 (2023)

\bibitem{Zhu_2018_apanet}
Zhu, Y., Wang, J., Xie, L., Zheng, L.: Attention-based pyramid aggregation network for visual place recognition. In: 2018 {ACM} Multimedia Conference on Multimedia Conference, {MM} 2018, Seoul, Republic of Korea, October 22-26, 2018. pp. 99--107. {ACM} (2018)

\end{thebibliography}
}
\end{document}